\documentclass{article}

\usepackage{arxiv}

\usepackage[utf8]{inputenc} 
\usepackage[T1]{fontenc}    
\usepackage{hyperref}       
\usepackage{url}            
\usepackage{booktabs}       
\usepackage{amsfonts}       
\usepackage{nicefrac}       
\usepackage{microtype}      
\usepackage{lipsum}		
\usepackage{graphicx}
\usepackage{natbib}
\usepackage{doi}

\usepackage{subcaption}
\usepackage{chngcntr}
\usepackage{caption} 
\title{A Deep Dive into Deep Cluster}

\author{{Ahmad Mustapha} \\
	Department of Electrical and Computer Engineering\\
	American University of Beirut\\
	Bliss Street, Beirut\\
	\texttt{amm90@mail.aub.edu} \\
	\And
	{Wael Khreich}\thanks{Corresponding Author} \\
	Olayan School of Business\\
	American University of Beirut\\
	Bliss Street, Beirut\\
	\texttt{wk47@aub.edu.lb} \\
	\And
	{Wassim Masri} \\
	Department of Electrical and Computer Engineering\\
	American University of Beirut\\
	Bliss Street, Beirut\\
	\texttt{wm13@aub.edu.lb} \\
}

\renewcommand{\shorttitle}{\textit{arXiv} Template}

\hypersetup{
pdftitle={A template for the arxiv style},
pdfsubject={q-bio.NC, q-bio.QM},
pdfauthor={David S.~Hippocampus, Elias D.~Striatum},
pdfkeywords={First keyword, Second keyword, More},
}

\begin{document}
\maketitle

\begin{abstract}
	Deep Learning has demonstrated a significant improvement against traditional machine learning approaches in different domains such as image and speech recognition. Their success on benchmark datasets is transferred to the real-world  through pretrained models by practitioners. Pretraining visual models using supervised learning requires a significant amount of expensive data annotation. To tackle this limitation, DeepCluster - a simple and scalable unsupervised pretraining of visual representations - has been proposed. However, the underlying work of the model is not yet well understood. In this paper, we analyze DeepCluster internals and exhaustively evaluate the impact of various hyperparameters over a wide range of values on three different datasets. Accordingly, we propose an explanation of why the algorithm works in practice. We also show that DeepCluster convergence and performance highly depend on the interplay between the quality of the randomly initialized filters of the convolutional layer and the selected number of clusters. Furthermore, we demonstrate that continuous clustering is not critical for DeepCluster convergence. Therefore, early stopping of the clustering phase will reduce the training time and allow the algorithm to scale to large datasets. Finally, we derive plausible hyperparameter selection criteria in a semi-supervised setting.
\end{abstract}


\section{Introduction}

Deep Learning has been topping benchmarks leader boards across different domains such as image recognition \citep{krizhevsky_imagenet_2017}, speech recognition \citep{amodei_deep_2015}, activity recognition \citep{wang_deep_2019}, and image generation \citep{karras_analyzing_2020}. In computer vision, this success has  mainly been attributed to ImageNet \citep{deng_imagenet_2009}. When an application has limited data, practitioners rely on models pretrained over ImageNet. 

However, pretrained models over ImageNet hold features that are limited to the classes present in the dataset and don't generalize to non-natural images - except for the lower layers that learn generic Gabor filters \citep{yosinski_how_2014}. Moreover, settling this problem by pretraining models on larger datasets is not efficient as labeled data is expensive, especially if domain expert annotators are needed. On the other hand, unlabeled data is cheap and freely available. Is it possible to have scalable unsupervised model pertaining approaches?

Unsupervised deep learning has been recently gaining researchers' attention. A recent approach is DeepCluster \citep{caron_deep_2019}, which is attractive for its simplicity and scalability. As far as we know, no one has tried to conduct a thorough analysis regarding why DeepCluster works and what is the effect of different hyperparameters on its performance. 

The contributions of this paper are the following. First, we provide useful and novel insights into the underlying mechanisms of DeepCluster and propose an explanation of why the algorithm works in practice. Our analysis and results are based on a comprehensive set of experiments conducted on three different datasets over a wide range of parameters (such as learning rate, weight decay, momentum, Sobel filter, batch normalization, number of cycles, number of clusters, dimensionality reduction, and training batch size). Second, we show that continuous clustering is not critical for the convergence of the DeepCluster algorithm, and stopping the resource-intensive clustering phase enormously reduces the training time of the algorithm and hence allows it to scale to large datasets. Third, we demonstrate that both convergence and performance of DeepCluster highly depend on the interplay between the quality of the randomly selected filters of the convolutional layer and the chosen number of clusters. Finally, we derive plausible hyperparameter selection criteria in a semi-supervised setting.

The rest of the paper is structured as follows. We present related work in section 2. In Section 3, we introduce Deep-Cluster approach. In section 4, we state our explanation of why Deep-Cluster works. In section 5, we study the effect of Deep-Cluster parameters on performance. In section 6, we conclude our paper.

\begin{figure}[t]
\begin{center}
   \includegraphics[width=0.65\linewidth]{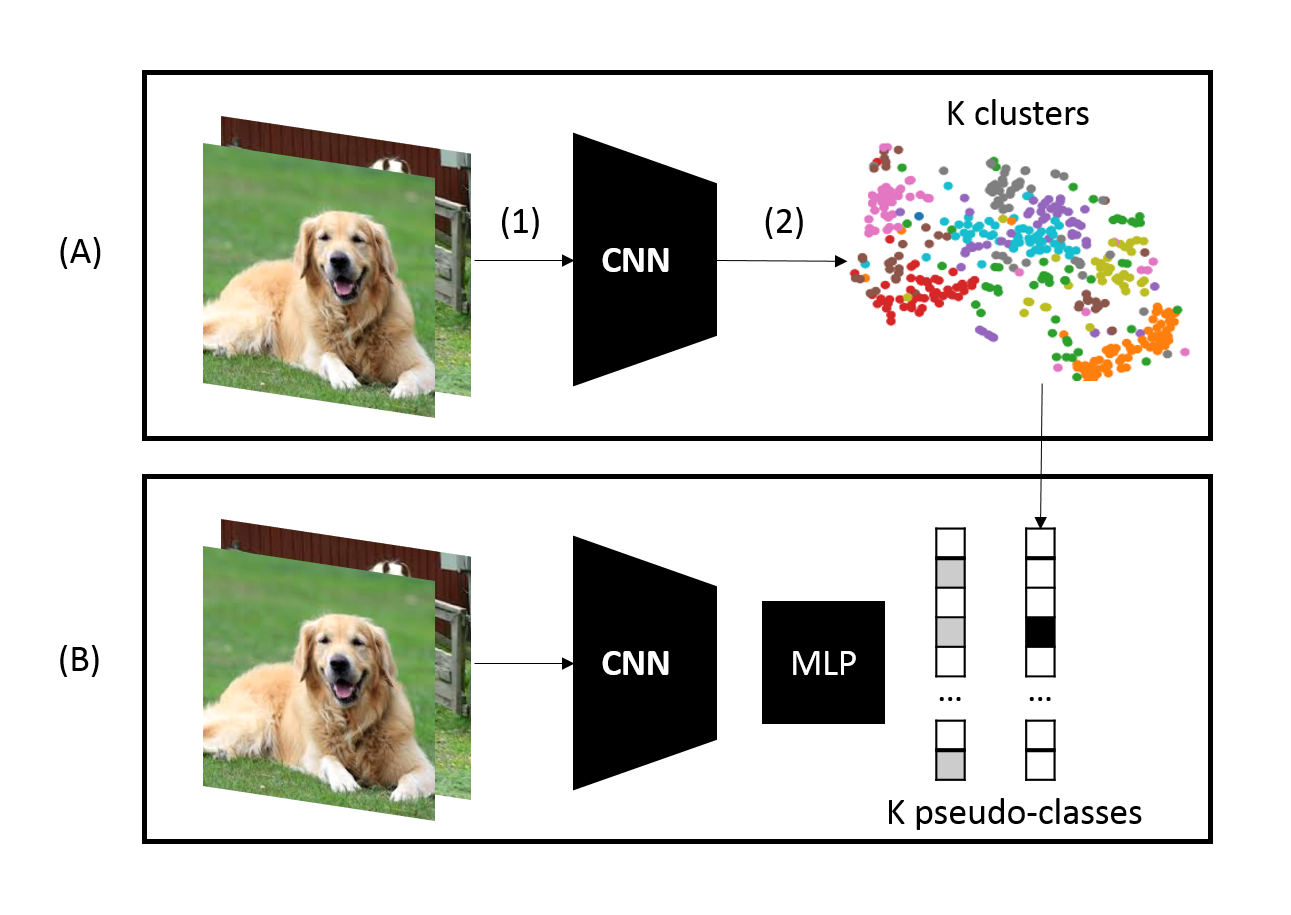}
\end{center}
   \caption{DeepCluster pipeline}
\label{fig:deepcluster}
\end{figure}

\section{Related Work}

Unsupervised Deep Representation Learning has existed since the early deep networks were proposed \citep{hinton_fast_2006, lee_convolutional_2009}. However, it has always been shadowed by the success of deep networks in supervised settings. In recent years, unsupervised representation learning has gained researchers' attention. 

One can divide deep representation learning approaches into a taxonomy of 5 categories. (1) Generative, (2) Self-supervised, (3) Clustering-based, (4) Sample Specificity, and (5) Contrastive. Note that those categories are not mutually exclusive and overlap with each other. 

\textbf{Generative}.Generative approaches make use of representations learned by networks that models input distribution whether through reconstruction loss mainly auto-encoders (AEs) \citep{vincent_extracting_2008} or through generating data that "fool" a discriminator network mainly Generative Adversarial Networks (GANs) \citep{donahue_adversarial_2017, donahue_large_2019}.

\textbf{Self-supervised}.Self-supervised approaches replace missing labels with a pretext task that can be formulated from the available unlabeled data, holding that training on the proposed task leads to general representations. Several tasks have been proposed such as in-painting patches \citep{pathak_context_2016}, solving a jigsaw puzzle \citep{noroozi_unsupervised_2017}, coloring \citep{larsson_learning_2017}, predicting cross channels \citep{zhang_split-brain_2017}. A complete survey is done in \citep{jing_self-supervised_2019}.  

\textbf{Clustering-based}. Clustering-based approaches extend conventional unsupervised learning algorithms with the representation learning powers of neural networks. The majority combines a network loss to learn representations and a clustering loss to force learned representations to be clustering-friendly. Earlier approaches focused more on the unsupervised learning part in which evaluation used traditional unsupervised learning metrics such as inertia, normalized mutual information, etc. A complete survey was done in \citep{aljalbout_clustering_2018}. Latest approaches \citep{caron_unsupervised_2020, ym_self-labelling_2019} considered representation learning including DeepCluster \citep{caron_deep_2019}.

\textbf{Sample-specificity}. Sample-specificity approaches resolve the problem of missing labels by considering each sample to be a class. In other words, networks are trained to discriminate each sample \citep{dosovitskiy_discriminative_2015, bojanowski_unsupervised_2017, wu_unsupervised_2018}. 

\textbf{Contrastive}. Contrastive representation learning is gaining attention and approaches from this class are leading the unsupervised representation learning benchmarks leaderboards \citep{chen_simple_2020, henaff_data-efficient_2020, li_prototypical_2020, he_momentum_2020}. In contrastive learning, multiple crops are deducted from an image and considered as a support for the image features. Those crops (positive samples) are contrasted with unrelated images (negative samples). The network task in this setting is to maximize the similarity between features of positive samples while maximizing the dissimilarity between features of positive and negative samples. 

\textbf{Analyzing Approaches}. Our work doesn't introduce new unsupervised representation learning approaches but analyses an existing one - DeepCluster. It complements and improves upon other work. \cite{asano_critical_2020} studied the effect of minimizing the dataset size and quality on the performance of different approaches and found that it is possible to maintain the same performance on the lower layers of the network while training on one rich image. \cite{caron_unsupervised_2019} studied the performance of DeepCluster itself on non-curated data. \cite{goyal_scaling_2019} studied the effect of scaling datasets on different approaches. \cite{kolesnikov_revisiting_2019} studied the effect of different choices considered in the learning pipeline like models and evaluation. As described in the following sections, we provide insights into the underlying mechanisms of DeepCluster and practical recommendations based on a comprehensive set of experiments conducted on three different datasets over a wide range of hyperparameters (from the learning rate to the training size).

\section{Background}
\subsection{DeepCluster}
DeepCluster is a simple algorithm for deep unsupervised representation learning, which is similar to supervised deep learning, but instead of using class labels - as they are unknown, the  algorithm iteratively finds and uses pseudo-labels.

Figure \ref{fig:deepcluster} introduces Deep Cluster, which mainly iterates between (A) generating pseudo-labels and  (B) learning the network parameters based on these pseudo-labels. We denote $N_c$ to be the number of cycles DeepCluster run and $N_e$ to be the number of epochs to train the network on pseudo-labels in each DeepCluster cycle. The original authors found $N_e=1$ to perform well in practice.

To generate pseudo-labels, the entire dataset is fed forward into the network and a feature vector is computed for each sample. In other words, the input is then projected from Euclidean space into a ConvNet feature space. Next, the vectors' dimensions are reduced using principal component analysis, whitened and L2-normalized. The KMeans clustering algorithm is then applied on generated features to assign similar samples into a cluster. The assignments of samples are considered to be the pseudo-labels on which the model will be trained. The reader is referred to the original paper \citep{caron_deep_2019} for more details.

\begin{figure}[t]
\begin{center}
   \includegraphics[width=0.65\linewidth]{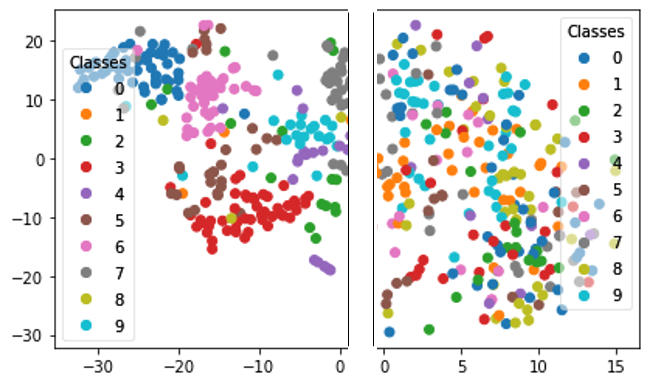}
\end{center}
   \caption{ (Left) MNIST-LeNet random feature space. (Right) CIFAR10-AlexNet random feature space }
\label{fig:random_filters_quality}
\end{figure}

\subsection{Evaluation Metrics}

\begin{figure*}[t]
    \begin{subfigure}[b]{0.3\textwidth}
          \includegraphics[width=\textwidth]{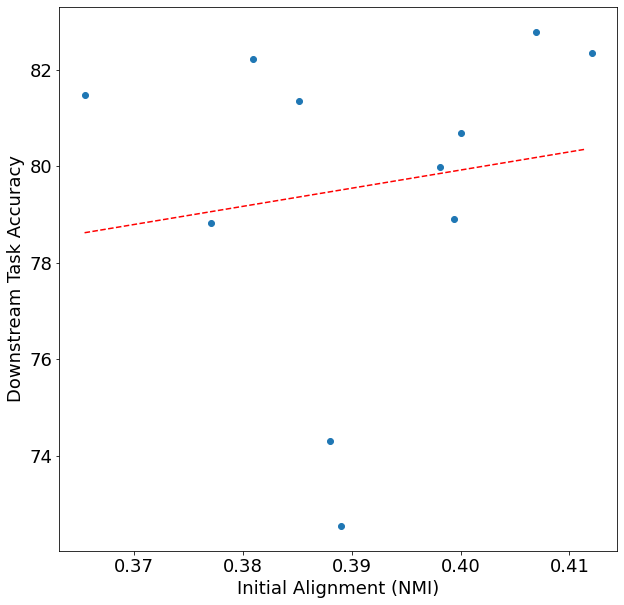}
          \caption{FMNIST}
    \end{subfigure}
    \hfill
    \begin{subfigure}[b]{0.3\textwidth}
        \includegraphics[width=\textwidth]{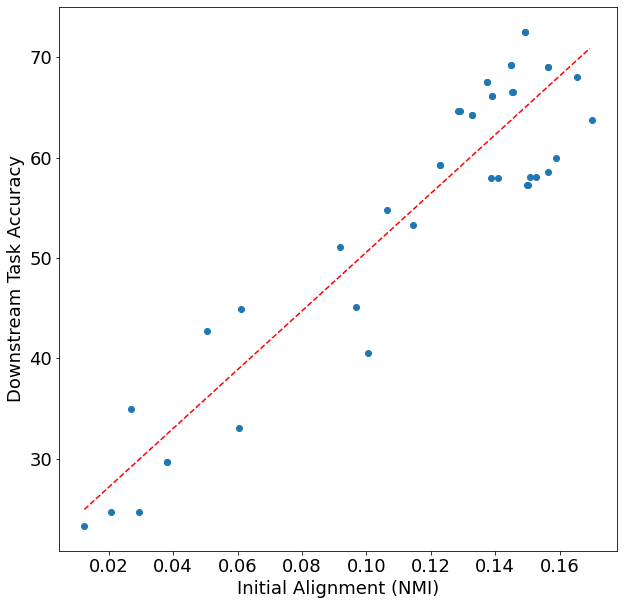}
        \caption{SVHN}
    \end{subfigure}
    \hfill
    \begin{subfigure}[b]{0.3\textwidth}
        \includegraphics[width=\textwidth]{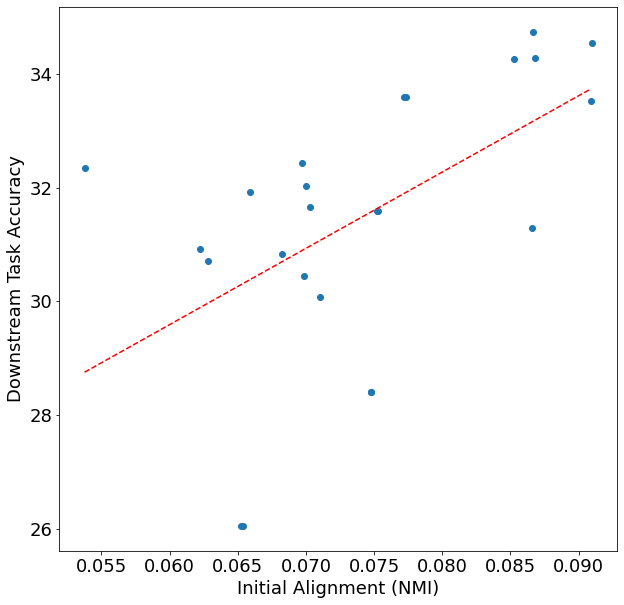}
        \caption{CIFAR10}
    \end{subfigure}
   \caption{Initial Alignment (IA) vs Downstream Task Accuracy}
\label{fig:change_theta}
\end{figure*}

Statistical learning-based computer vision requires a mapping of images from their raw space into a space that can capture image semantics. Convolutional networks are found to be useful for this type of mapping and outperform other approaches. Let $f_\theta$ be a ConvNet that maps images with $\theta$ being its parameters. The goal of training such networks is to find $\theta$ that makes $f_\theta$ generate decent general-purpose features. Given a set of $n$ images $X=\{x_0, x_1, ..., x_n\}$ with a label $y_i$ for each image $x_i$. Let $Y = y_i$ for $i \in \{0, 1, ..., n\}$ . The network is trained by predicting labels using a parameterized classifier $g_w$ over $f_\theta$. The parameters $\theta$ and $W$ are jointly learned through a loss function.

In DeepCluster the labels are not available but presumed by clustering the ConvNet feature space. In other words for each $x_i$ we compute $v_i=f_\theta(x_i)$, and the clustering is performed on $V = v_i$ for $i\in \{0, 1, .., n\}$. For each $x_i$ we generate a pseudo-label $y^p_i$ being the cluster assignment of the sample. Let $Y^p = y^p_i$ for each $i\in \{0, 1, .., n\}$. For each DeepCluster cycle $c$ a pseudo-label vector $Y^p_c$ is generated and the network is trained upon it.

To study how the learned feature space quality evolves during training, two metrics have been proposed:
\begin{enumerate}
\item $NMI(Y^p_{c_-1}, Y^p_c)$  which is the normalized mutual information between the cluster assignments of two consecutive DeepCluster cycles $c$ and $c-1$. NMI is a value that is bounded by $[0, 1]$ and represents the dependence between two groupings, zero means no dependence and one means that both groupings hold the same information and we can predict one from the other. If the model converges then $NMI(Y^p_{c_-1}, Y^p_c)$ should increase and become constant more or less at some point of DeepCluster training. 
\item $NMI(Y^p_c, Y)$ which computes the dependence between DeepCluster groupings at cycle $c$ and the ground truth grouping of samples. If the model is actually learning representative features then this metric should increase and become stable at some point of training.  
\end{enumerate}

While the above metrics are used as a preliminary convergence test they aren't used to evaluate the model. To evaluate unsupervised representation learning the community has developed the following framework. After unsupervised pretraining of a  model, the learned features are evaluated against a downstream task. Different downstream tasks are being used including classification. For classification the procedure goes as follows: the model feature extraction part $f_\theta$ is frozen and a linear classifier is trained in a supervised manner over each convolutional layer of the model and accuracy on each layer is reported. This process is called linear probing \citep{zhang_split-brain_2017}.

\subsection{Implementation Details}
\textbf{Training Data.} We experimented with the following three datasets ordered by increasing complexity FMNIST \citep{xiao2017/online}, SVHN \citep{svhn}, and CIFAR10 \citep{krizhevsky_learning_nodate} which are packaged in PyTorch \citep{NEURIPS2019_9015}. We used the official train/test split. For CIFAR10 we resized images into 32x32 ones. For the clustering part, we used the train set.

\textbf{Networks Architecture.} For FMNIST we used the LeNet5 \citep{lecun_gradient-based_1998} architecture. For SVHN and CIFAR10 we used a minimized variant of AlexNet \citep{krizhevsky2012imagenet} with five convolutional layers with 48, 126, 192, 192, and 128 filters respectively. We used Batch Normalization for all networks. We also used a Sobel filter that precedes the network in the case of CIFAR10 and SVHN datasets.

\textbf{Training Optimization.}
We selected the baseline training hyperparameters based on grid search. Table \ref{table:grid_search} shows the parameters searched along with the best-performing ones. Note that for CIFAR10 we only optimized the most important hyperparameters (which have the largest impact on performance). For all experiments, we used Stochastic Gradient Decent (SGD) optimizer. We set the learning rate to 0.1 and the weight decay to 0.001. We set the momentum for FMNIST, SVHN, and CIFAR10 to 0.1, 0.5, and 0.9 respectively. For SVHN and CIFAR10 we used a batch size of 64 and of size 128 for FMNIST. During the clustering phase, the transformations used are normalization, resizing, and then taking a center crop of size 32. While in the training phase we also used horizontal flip. For the dimensionality reduction step after clustering, we used principal component analysis (PCA) with 256 components for SVHN, and CIFAR10.

\begin{table*}[]
    \centering
    \begin{tabular}{c|c|c|c}
        Hyperparameter &  FMNIST & SVHN & CIFAR10\\
        \hline
        Learning rate & \textbf{[0.1}, 0.01, 0.001, 0.0001] & [\textbf{0.1}, 0.2, 0.01, 0.001] & \textbf{0.1} \\
        Weight Decay & [\textbf{0.001}, 0.0001, 0.00001] & [\textbf{0.001}, 0.01, 0.1] & \textbf{0.001} \\
        Momentum & [\textbf{0.1}, 0.5, 0.9] & [0.1, \textbf{0.5}, 0.9] & \textbf{0.9} \\
        Sobel Filter & \textbf{Off} & \textbf{On} & \textbf{On} \\
        Batch Normalization & [\textbf{On}, Off] & \textbf{On} & \textbf{On} \\
        Number of cycles & \textbf{50} & \textbf{100} & \textbf{100} \\
        Number of clusters & [\textbf{5}, 10, 20, 200] & [5, 10, 20, \textbf{200}, 1000] & [5, 10, 200, \textbf{1000}, 2000, 5000] \\
        PCA & [\textbf{Off}, 40, 60] & [Off, 64, 128, \textbf{256}] & [Off, 64, 128, \textbf{256}] \\
        Training batch size & [64, \textbf{128}, 256] & [\textbf{64}, 128, 256] & \textbf{64} \\
    \end{tabular}
    \caption{Grid search used for finding best performing hyperparameters. Selected ones are in bold.}
    \label{table:grid_search}
\end{table*}

\textbf{Training and Evaluation Procedure.} We run DeepCluster for 50 cycles over FMNIST and 100 cycles over SVHN and CIFAR10. In each cycle, we do (1) a clustering step including a full feed-forward over the training set followed by clustering, whitening, L2 normalization, and PCA and (2) a training step of one epoch. Finally, to evaluate the process we freeze the network and train a logistic regression classifier over the topmost layer (ReLu-5 for SVHN and CIFAR10 and Conv-2 for FMNIST ) for 20 epochs and then use the test dataset for evaluation.  

\section{Experiments}
\subsection{Why DeepCluster Works}

As far as we know no one has tried to understand why such an approach works out. Intuitively DeepCluster groups samples in proximity together and train the model to discriminate between different groups. According to the authors, DeepCluster makes use of the original random features - the randomly initialized filters of the convolutional layers - which gives a strong prior to the input signal due to their convolutional structure \citep{saxe_random_nodate}. It makes sense to think less of the randomly initialized weights of a convolutional layer as numerical matrices, and think more of them as random convolutional filters that hold loose spatial signals.

This raises the question of how much the random filter quality affects the final results of DeepCluster on downstream tasks or on its representation learning capabilities and what role the number of clusters plays in this process. In this manuscript, we argue that DeepCluster success strongly depends on both the quality of the random filters and the choice of the number of clusters. The more the random filters are suitable for the images domain, the more projected samples in proximity hold similar semantic cues. The more clusters we use the more feature consistent clusters we get until reaching a point where over segmentation degrades performance.

Consider the following illustrative example. (A) We feed a sample of MNIST dataset into a randomly initialized LeNet network. (B) We feed a sample of CIFAR10 dataset into a randomly initialized AlexNet network. In Figure \ref{fig:random_filters_quality} we plot the final convolutional layer values for both networks reduced using t-SNE \citep{van2008visualizing} along with the ground truth as colors. The random filters are suitable for the digit images domain. In other words, projecting the images through the random filters leads to a space where similar digits fell in the same regions. However, this is not the case for CIFAR10, as similar classes of images don't fall in wide regions. Nonetheless, we can see a large sum of them that fell near each other in narrow regions. To have clusters of the same class we have to increase the number of clusters. The number of clusters chosen can then resolve the weakness in the random filter suitability for CIFAR10 classes.

This proposal explains why the original authors found that DeepCluster performed better on downstream tasks when they chose $10,000$ clusters than when they chose $1,000$ clusters over the ImageNet dataset (which contains $1,000$ classes). This is also explains why transforming images using the Sobel filter performed better (random filters are more suitable for discriminating shapes but not patterns). We further support this claim with a set of experiments in the following subsections.

\subsubsection{Random Filter Suitability Metric}

To demonstrate DeepCluster dependence on the suitability of random filter and the selected number of clusters, we formulated the following metric. Let $K$ be the number of clusters used in each DeepCluster cycle. Let $\theta$ be the random filters generated given a random seed and used as a starting point for the network training. 

We hypothesize that the pair $(k, \theta)$, decides together the performance of DeepCluster. To measure the quality of this pair, or how much the pair is suitable for the dataset at hand, we define the metric $IA(k, \theta)$.

Let $L$ be the class labels of the dataset samples - $L$ is not used during DeepCluster iterations but we use it here for experimenting purposes. 
Let $P$ be the pseudo-labels (i.e. clustering assignments) obtained by applying KMeans using $K$ clusters over the feature space produced by projecting $X$ through $f_\theta$. Note that $L$ takes values between 0 and $C$ where $C$ is the number of classes in the dataset while $P$
takes values between $0$ and $K$. Then $IA(\theta, K)$ is defined as:
$$IA(\theta,k) = NMI(L,P)$$
where NMI is the Normalized Mutual Information.

The higher this metric is, the more the choice of $K$ complements the suitability of $\theta$ to have pseudo-classes that align with the ground-truth classes. Back to the illustrative example in Figure \ref{fig:random_filters_quality}, because the random filters are not very suitable for CIFAR10 we need to increase the number of clusters. However, we need to use a smaller number of clusters for MNIST, because the random filter suites the data. To test how much DeepCluster depends on this metric, we conducted two experiments. In the first, we varied the randomization seed. In the second, we varied the number of clusters. We will call this metric Initial Alignment (IA).

\begin{figure}[t]
\begin{center}
   \includegraphics[width=0.65\linewidth]{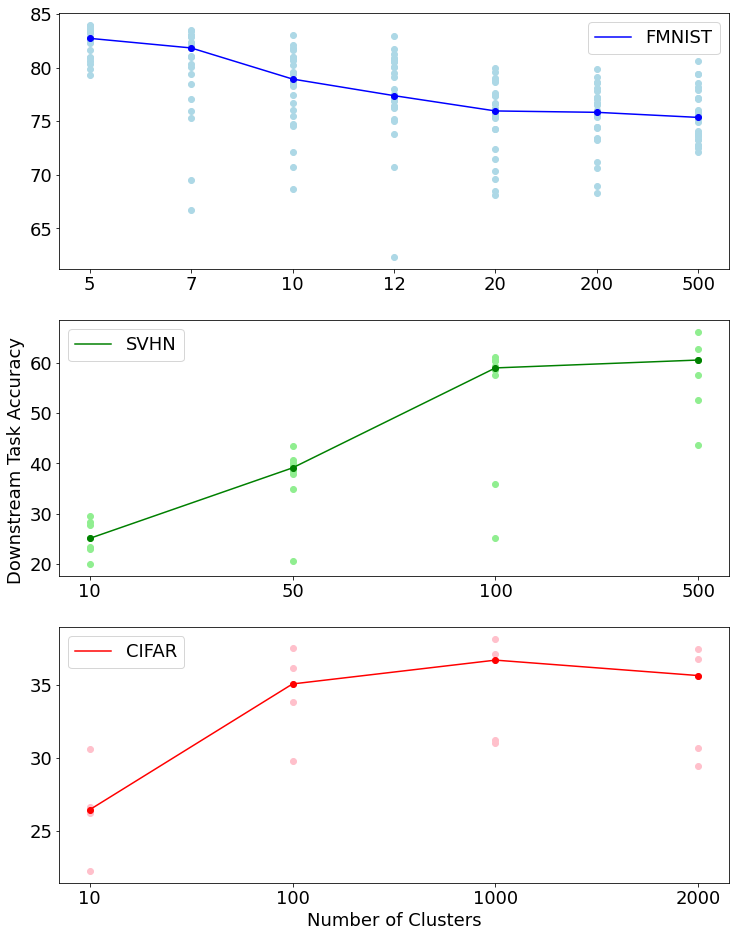}
\end{center}
   \caption{Number of Clusters vs. Downstream Accuracy}
\label{fig:change_k}
\end{figure}

\subsubsection{Effects of random filters suitability to data domain}

In this experiment we fixed the number of clusters based on grid search and only varied the randomization by changing the random state. We then computed IA in each case and plotted them against the evaluation accuracy. In other words, we are varying $\theta$ and fixing $K$ while computing $IA(\theta, K)$.

We selected the seeds such that they span the entire range of $IA(\theta, K)$ distribution. To sample from the distribution we ran $100$ full feed-forward steps followed by a clustering step and reported $IA$. The results are reported in Figure \ref{fig:change_theta}.

First we can see that the range of $IA$ of FMNIST dataset is high despite K=5 (0.37~0.41) (Figure \ref{fig:change_theta}a) while it is smaller for SVHN (0.04~0.16) with $K=200$ (Figure \ref{fig:change_theta}b) and much smaller for CIFAR10 (0.065~0.085) with $k=1,000$ (Figure \ref{fig:change_theta}c). This hints at the suitability factor we mentioned above. FMNIST is characterized by very simple features that if projected through random filters will result in a space that reflects the different classes. The suitability decreases as we increase the complexity of the dataset features as in the case of SVHN and CIFAR10. Second, we see a positive relationship between IA and accuracy on the downstream task in the case of SVHN and CIFAR10 but not of FMNIST. This clearly shows that the quality of the random filter affects the downstream task accuracy. This pattern doesn't affect FMNIST as the lowest IA value is good enough to bootstrap learning. 

\subsubsection{Effects of clusters numbers}

In this experiment, we fixed a set of seeds and varied the number of clusters. The results are reported in Figure \ref{fig:change_k}.

For FMNIST we have a negative relationship between the number of clusters and accuracy (Figure \ref{fig:change_k}a). This again points out the fact that when random filters are suitable for the dataset domain, a small number of clusters suffices for DeepCluster to converge and perform well. In fact increasing the number of clusters will decrease accuracy because the model will be learning to separate images that hold very similar semantic, which leads it to learn subtle details and decreases its generalization.

On the other hand, on SVHN and CIFAR10 (Figure \ref{fig:change_k}b and \ref{fig:change_k}c respectively) the number of clusters has a positive relationship with the downstream task accuracy. This can be explained by the fact that random filters represent a very weak spatial signal. This signal is not enough to capture complex features such as animal faces. As a result, when using random filters as a projection layer they lead to a space where only very similar images (from a shape perspective) appear in proximity. If we used a small number of clusters for such image space, we will end up having clusters that don't contain common patterns. For example, we might get frogs and horses in the same cluster. Training on these clusters will lead the model to learn noise and non-generalizable features. On the other hand, increasing the number of clusters will lead to more consistent features. For an illustrative example refer to CIFAR10 space under random filters in Figure \ref{fig:random_filters_quality}.

\subsection{Is continuous clustering important?}

One of the most important questions related to DeepCluster is whether clustering should be performed throughout the cycles or not. While the first clustering operation is integral, there is no empirical evidence on the importance of continuous clustering. In other words, how will the accuracy on the downstream task be affected if we only cluster once and then use the pseudo-labels to train the network in a supervised fashion over them? To answer this question we ran a set of experiments where everything is fixed except for stopping clustering at different cycles. 

\begin{figure*}[t]

    \begin{subfigure}[b]{0.3\textwidth}
          \includegraphics[width=\textwidth]{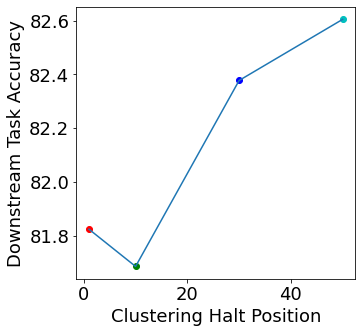}
          \caption{FMNIST}
    \end{subfigure}
    \hfill
    \begin{subfigure}[b]{0.3\textwidth}
          \includegraphics[width=\textwidth]{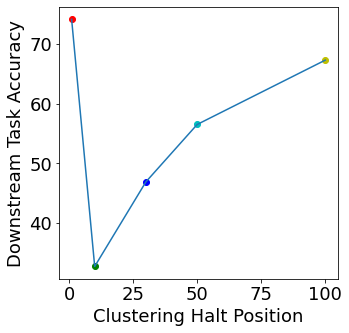}
          \caption{SVHN}
    \end{subfigure}
    \hfill
    \begin{subfigure}[b]{0.3\textwidth}
          \includegraphics[width=\textwidth]{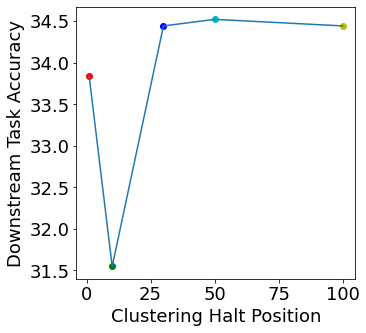}
          \caption{CIFAR10}
    \end{subfigure}
    
   \caption{The effect of halting clustering after a certain number of cycles over downstream task accuracy (accuracy)}
\label{fig:change_halt}
\end{figure*}

To make this clearer, assume we are performing 100 DeepCluster cycles in each cycle we have one clustering operation and one training operation. If we halt clustering at cycle 50 then we will be performing 50 cycles with clustering followed by 50 training-only cycles using the latest pseudo-labels of the 50th cycle. The result of the experiments are reported in Figure \ref{fig:change_halt}.

For both FMNIST and CIFAR10, continuous  clustering slightly improved the accuracy, approximately one degree in accuracy. But for SVHN there is an effect of approximately 40 degrees in accuracy. Interestingly depending only on the first clustering operation over SVHN performed better than keeping on clustering for 10, 30, 50, and 100 cycles! Another consistent observation across the three datasets is that accuracy decreases as we perform clustering before it increases again.

This points out to the fact that continuous clustering is not necessarily important for performance on downstream tasks. Considering that DeepCluster is supposed to be run on datasets of the scale of ImageNet  and more and that each clustering step is performed after feeding the entire dataset into the network, this experiment shows that we can trade-off a considerable computational load with limited or no degradation in accuracy.

\subsection{Initial Alignment of Hyperparamter Selection}

Given that we empirically established a relationship between the IA score and  the overall accuracy of downstream tasks, in this section we investigate if we can use IA as a heuristic for choosing hyperparameters. For example rather than running DeepCluster for different number of clusters to decide the best value, as the authors did in the original paper, we can only compute IA and use it to guide the selection process.


The hypothesis is that computing the different IA distributions resulting from different values for a given hyperparameter, will allow us to choose the best choice by selecting  the value that results in the highest IA score. 

As the IA score depends on a random variable we have to sample the IA distribution resulting from a given hyperparameter value by computing IA multiple times while changing randomization seed. 

To study this possibility we sampled the distribution of IA over SVHN for the following values of the number of clusters [10, 50, 100, 500] and then we take the IA median of the distribution (check appendix \ref{app:moreOnIA}.

To sample the distribution we fixed all the hyperparameters including the targeted hyperparameter which is the number of clusters in this case and then we (1) feed the entire dataset throughout the randomly initialized projection layer (2) compute KMeans on the resulting projected image vectors (3) compute the IA. We repeated this  for $100$ times while changing only the random state which leads to different initialization of weights every time. For each sampled IA distribution we note the median value.

\begin{figure}[t]
\begin{center}
   \includegraphics[width=0.65\linewidth]{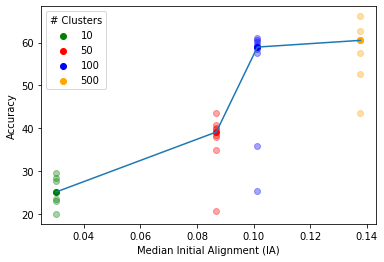}
\end{center}
   \caption{Initial Alignment IA can be used to guide hyperparameters selection. Here we see the relation between IA distribution under the chosen number of clusters and downstream task accuracy.}
\label{fig:vary_k_svhn_space_dist_perf}
\end{figure}

After that we sample the downstream task accuracy distribution for each number of clusters by running DeepCluster for several times and noting the median accuracy. Using SVHN we sampled accuracy distribution using five runs (see Section \ref{sec:limitation}. 
The results can be seen in Figure \ref{fig:vary_k_svhn_space_dist_perf}.

As seen in Figure \ref{fig:vary_k_svhn_space_dist_perf}, the more the IA distribution median have a higher value, the more the downstream task median accuracy is. Choosing 10 or 50 as a number of clusters results in an IA median value that is lower than 0.08 and accuracy ranging between [20 and 40]. On the other hand, choosing 100 or 500 as a number of clusters results in an IA median value that is greater than 0.1 and accuracy ranging between [25 and 60].

Moreover, the difference between the median IA of 100 clusters and 500 clusters is of 4 degrees while the difference between the median accuracy of both is less than 2 degrees. This means that after a certain IA median value accuracy starts to saturate. 

This experiment shows that IA distribution can be used as a hyperparameter selection guidance, which can help selecting the best hyperparameters and decreasing the optimization scope.
For example, for SVHN we can discard 10 and 50 as a value for the number of clusters because of their relatively low IA distribution median. 

Note that in unsupervised learning settings it is not possible to compute the IA because it is based on ground truth labels. This application of IA is only possible in a semi-supervised setting where there is a sufficient number of ground truth labels.

\section{Limitation And Future Work}
\label{sec:limitation}

Our experiments were performed on medium-sized datasets and using small architectures. For CIFAR10 in particular we used a smaller version of AlexNet. This has degraded the overall downstream task accuracy as the dataset requires models with higher capacity. Initially, we used the original AlexNet (check appendix \ref{app:alexnet} but because of the prohibitive computational power requirements, we were forced to use a smaller architecture given that our goal is to study the relationship between the IA and downstream task performance.

We have shown that the number of clusters and random filter quality have large impact on the DeepCluster performance. Thus it is interesting to investigate different procedures that can enhance those two hyperparmaters. From one side rather than choosing a large number of clusters we can choose a smaller number by  formulating the problem as a Learning with Noisy Labels (LNL) \citep{han_co-teaching_2018} to reduce the impact of inconsistency on learning. On the other side, it is possible to raise the quality of random filters either by applying sample-specificity \citep{wu_unsupervised_2018} pretraining as in \citep{huang_unsupervised_2019} or by developing better transformations on the images such as making use of a ScatterNet \citep{bruna_invariant_2012} or even consider starting out from pretrained networks.

Finally, we think that IA usability can be even extended into a supervised learning setting. As it can be used as a hyperparameter selection and as a driver for network architecture search (NAS) in order to find architectures that are best suited for the input domain.

\section{Conclusion}

We proposed an explanation of why the algorithm works in practice supported by a comprehensive set of experiments over three different datasets. First, we found that DeepCluster convergence depends on two key hyperparameters, the initial random filters and the number of clusters. These hyperparameters have a large impact on the overall accuracy of the representation learned by DeepCluster as measured on various downstream tasks. We used these insights to propose a parameter selection approach that proved effective and certainly better than an exhaustive search over the entire range of hyperparameters. We then investigated the contribution of continuous clustering on DeepCluster accuracy and found that clustering acts simply as a space partition, and its contribution is minimal. Therefore, early stopping of the clustering phase makes the algorithm more scalable for large datasets.

\bibliographystyle{unsrtnat}
\bibliography{references}

\newpage

\section*{Appendix}
\renewcommand{\thesection}{\Alph{section}}
\setcounter{section}{0}
\counterwithin{figure}{section}
\counterwithin{table}{section}

\section{Experiments over CIFAR10 using AlexNet}\label{app:alexnet}
Initially we applied our experiments on CIFAR10 while using an AlexNet architecture. However, when we find that we will need to repeat experiments multiple times we switched to a smaller variant of AlexNet. In this appendix section we report different experiments results that we get over CIFAR10 using AlexNet. For the below experiments we set learning rate to 0.05, momentum to 0.9, weight decay to 0.00001, number of clusters to 2000, number of PCA components to 256, batch size to 256, we used batch normalization and sobel filters unless stated otherwise.

\subsection{IA vs Downstream Task Accuracy}
\begin{figure}[h]
    \centering
    \includegraphics[width=0.4\textwidth]{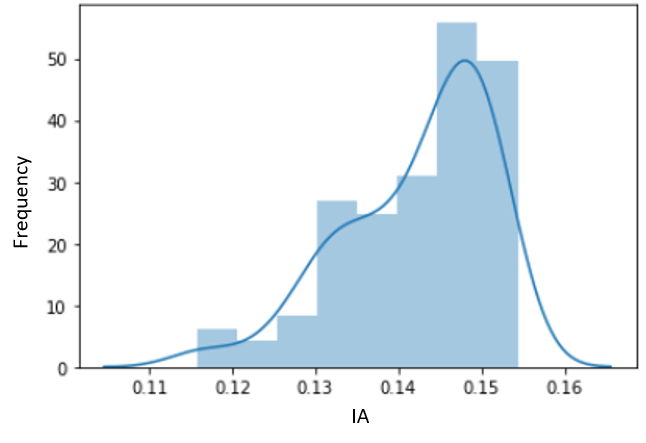}
    \includegraphics[width=0.42\textwidth]{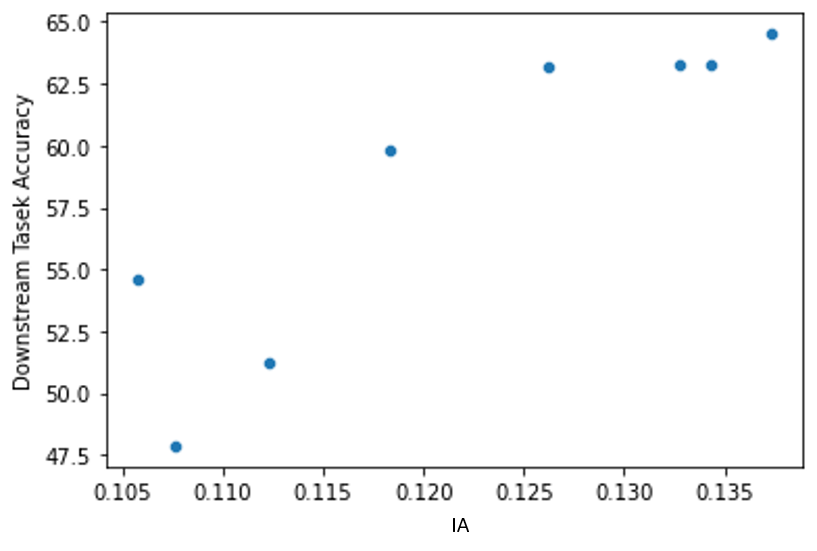}
    \caption{(Left) Distribution of IA over CIFAR10 using AlexNet with k=2000. (Right) IA vs Downstream Task Accuracy.}
    \label{fig:change_theta_dist_appendix}
\end{figure}

To quantify the relation between IA and downstream task accuracy we first sampled the IA distribution then noted 8 random states that corresponds to different IA values. After that we run DeepCluster for each random state. To sample IA distribution we performed 100 runs while computing IA in each run. The resulting distribution and the corresponding accuracy of the selected random states is reported in Figure \ref{fig:change_theta_dist_appendix}. The results align with the result of running the experiment over SVHN and over CIFAR10 with a smaller AlexNet variant: the higher IA is the higher is the accuracy on the downstream task.

\subsection{Continuous Clustering Contribution}
In this experiment We quantified the effect of continuous clustering over CIFAR10 using AlexNet. To quantify this we performed several DeepCluster runs while having all parameters fixed including random state except for halting clustering at different positions. We halted clustering after 15, 30, 50, and 100 cycles. The results are reported in the following table \ref{table:cont_clustering_appendix}.

\begin{table}[h]
    \centering
    \begin{tabular}{c|c}
    \hline
         Halt Clustering Position &  Downstream Task Accuracy \\
    \hline
         15  &	62.00 \\
         30  &	64.06 \\
         50  &	65.18 \\
        100  &	66.98 \\
        \hline
    \end{tabular}
    \caption{Continuous clustering contribution to downstream task accuracy over CIFAR10 using AlexNet.}
    \label{table:cont_clustering_appendix}
\end{table}

The results align with the ones reported for SVHN and CIFAR10 using the smaller variant of AlexNet: Continuous clustering acts as a regularizer and it's effect on the downstream task accuracy is minimal.

\subsection{PCA Components Contribution}
In this experiment we quantified the effect of choosing the number of PCA components over the downstream task accuracy. We simply performed several DeepCluster runs while fixing all parameters including random state except for the number of PCA components. The results are reported in the following table \ref{table:pca_components_contribution}.

\begin{table}[h]
    \centering
    \begin{tabular}{c|c}
    \hline
    PCA &  Downstream Task Accuracy \\
    \hline
    32 	& 64.97 \\
    64 	& 64.80 \\
    128 & 65.50 \\
	256 & 65.47 \\
	\hline
    \end{tabular}
    \caption{PCA components contribution to downstream task accuracy over CIFAR10 using AlexNet.}
    \label{table:pca_components_contribution}
\end{table}

The results hints to the minimal contribution that the number of PCA components play in the downstream task accuracy. This means that it is possible to trade-off small accuracy degradation in return for significant computational power reduction. 

\section{IA For Hyperparameter Selection} \label{app:moreOnIA}
This section supplements the paper with more figures regarding using IA as a heuristic for hyperparameters selection. 

\subsection{IA Distribution Over SVHN}
\begin{figure}[h]
    \centering
    \includegraphics[width=\textwidth]{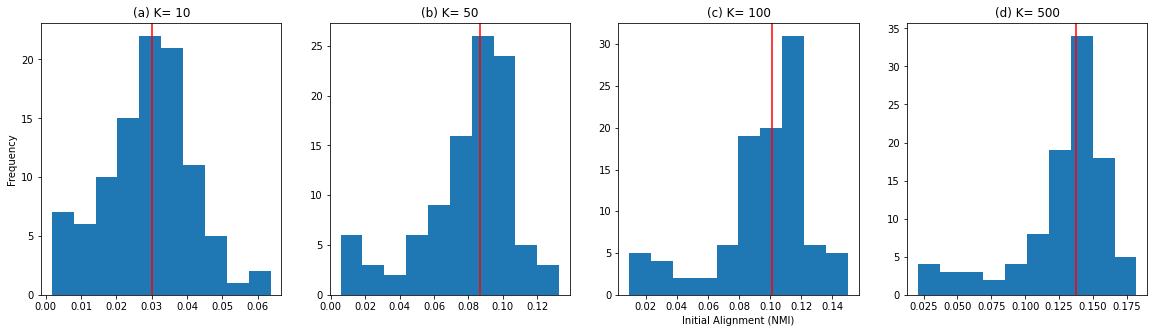}
    \caption{IA distribution for different number of clusters over SVHN dataset. The median of each distribution is marked by a red vertical line.}
    \label{fig:svhn_clusters_ia_dist}
\end{figure}

In Figure \ref{fig:svhn_clusters_ia_dist} we plot the sampled IA distribution over SVHN dataset for different values for the number of clusters [10, 50, 100, and 500]. The figure shows how increasing the number of clusters results in an IA distribution that is skewed toward higher IA values.

\subsection{IA For Selecting PCA Components}
\begin{figure}[h]
    \centering
    \includegraphics[width=\textwidth]{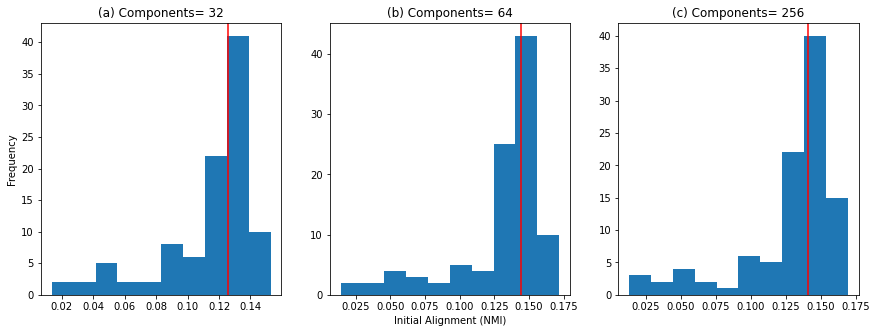}
    \caption{IA distribution for different number of PCA components over SVHN dataset. The median of each distribution is marked by a red vertical line.}
    \label{fig:svhn_pca_ia_dist}
\end{figure}

\begin{figure}[h]
    \centering
    \includegraphics[width=0.6\textwidth]{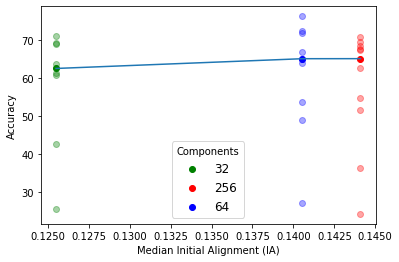}
    \caption{The relation between IA distribution of different number of PCA components and downstream task accuracy.}
    \label{fig:svhn_pca_ia_acc_dist}
\end{figure}

In this section we performed another experiment regarding selecting hyperparameters based on IA distributions. The targeted hyperparameter in this experiment is the number of PCA components. We fist sampled IA distribution for each candidate value of number of components [32, 64, 256] by performing 100 runs and computing IA in each run. The distributions are presented in Figure \ref{fig:svhn_pca_ia_dist}. All the three different values of PCA results in a very similar distribution that is skewed toward high values of IA. In other words the number of PCA components doesn't effect the initial alignment values. If IA can be used to guide hyperparameter selection then we expect do have similar downstream task accuracy for all the three number of PCA components. To do check that we run DeepCluster for several times for each candidate value of number of clusters. In Figure \ref{fig:svhn_pca_ia_acc_dist} we plot the median values of IA distributions vs accuracy. All three candidate values have a similar accuracy distribution. The figure shows that having similar IA distributions will lead to similar accuracy distribution. This again supports the fact that IA can be indeed used as a heuristic for hyperparameters selection.

\end{document}